\documentclass[conference]{IEEEtran}
\IEEEoverridecommandlockouts

\usepackage{amsmath,amssymb,amsfonts}
\usepackage{algorithmic}
\usepackage{graphicx}
\usepackage{textcomp}
\usepackage{optidef}
\usepackage{xcolor}
\usepackage{amsmath}
\usepackage{siunitx}
\usepackage{booktabs}
\usepackage{hyperref}
\usepackage[utf8]{inputenc}
\usepackage{cite}

\def\BibTeX{{\rm B\kern-.05em{\sc i\kern-.025em b}\kern-.08em
    T\kern-.1667em\lower.7ex\hbox{E}\kern-.125emX}}
\usepackage[acronym]{glossaries}
\glsdisablehyper

\newacronym{mpc}{MPC}{model predictive control}
\newacronym{rl}{RL}{reinforcement learning}
\newacronym{wb}{WB}{whole-body}
\newacronym{srb}{SRB}{single-rigid-body}
\newacronym{com}{COM}{center-of-mass}
\newacronym{wbc}{WBC}{Whole-Body Control}
\newacronym{ddp}{DDP}{differential dynamic programming}
\newacronym{ilqr}{ILQR}{iterative linear quadratic regulator}
\newacronym{sqp}{SQP}{sequential quadratic programming}
\newacronym{dof}{DoF}{degrees-of-freedom}
\newacronym{nlp}{NLP}{non-linear program}
\newacronym{ipm}{IPM}{interior-point method}
\newacronym{qp}{QP}{quadratic program}
\newacronym{admm}{ADMM}{alternating direction method of multipliers}
\newacronym{ocp}{OCP}{optimal control problem}
\glsunset{dof}

\usepackage{longtable}

\begin{document}

\title{Right Model, Right Time: Real-Time Cascaded-Fidelity MPC for Bipedal Walking\\
}
\author{Franek Stark$^{1,2}$, Felix Wiebe$^{1,2}$, Shubham Vyas$^{1,2}$, Dennis Mronga$^{1}$, Frank Kirchner$^{1,2}$
\thanks{$^{1}$All authors are with the Robotics Innovation Center at the German Research Center for Artificial Intelligence (DFKI), Bremen, Germany}%
\thanks{$^{2}$FS, FW, SV, and FK are also with the University Bremen, Germany}%
\thanks{Corresponding author's email: {\tt\footnotesize franek.stark@dfki.de}}
}

\maketitle

\begin{abstract}
This paper presents a multi-phase whole-body model predictive control (MPC) approach for bipedal walking, combining a detailed whole-body model in the near horizon with a simplified single-rigid-body model in the later prediction steps. This reduces computational complexity while retaining prediction capabilities.
The resulting nonlinear optimal control problem is solved entirely within the general-purpose, off-the-shelf nonlinear MPC framework acados, using sequential quadratic programming (SQP).
Given a contact schedule and a target walking speed, the controller optimizes joint torques without depending on pre-selected footstep locations.
The controller is validated in MuJoCo simulation on the 18-DoF bipedal robot HyPer-2.
\end{abstract}


\section{Introduction}
Legged robotics has advanced rapidly in recent years, with dynamic locomotion successfully demonstrated on quadrupeds and, later, on bipedal robots. 
Thereby, two different control paradigms are prevalent: \gls{rl} and \gls{mpc}. While \gls{rl} has shown impressive results and benefits strongly from improved simulations and computing power, it still faces well-known challenges such as time-consuming reward shaping, sample inefficiency, lack of generalization and open questions regarding safety. For this reason, \gls{mpc} remains attractive, because it allows the controller to exploit system dynamics online and reason about constraints directly.
However, for high-dimensional problems, such as bipedal walking in humanoids, the computational complexity of the optimal control problem remains too high, especially when aiming for an online solution. 
Hence, many works rely on reduced-order models in \gls{mpc}, in combination with instantaneous \gls{wbc}, which uses the full dynamics \cite{kim_highly_2019}. A common limitation of these approaches is that contact forces and body motions are  planned independently. Here, additional heuristics make systematic tuning of the different control levels difficult and the use of simplified models in \gls{mpc} may lead to dynamically infeasible motion plans.
Recently, whole-body \gls{mpc} for bipedal robots has become feasible due to advances in fast numerical optimal control, demonstrated with \gls{ddp}-based solvers~\cite{dantec_whole-body_2022} and \gls{sqp}-based approaches~\cite{galliker_planar_2022, khazoom_tailoring_2024}.
These approaches use a single full-fidelity model throughout the prediction horizon, which limits horizon length under real-time constraints.
Multi-phase (cascaded-fidelity) \gls{mpc} formulations address this by using a detailed model in the early steps of the prediction horizon and simpler models in later steps~\cite{li_model_2021, li_cafe-mpc_2025,khazoom_tailoring_2025}, reducing problem complexity while preserving predictive capability.
These works rely on problem-specific \gls{ddp} solvers~\cite{li_model_2021, li_cafe-mpc_2025} or a custom \gls{sqp} solver~\cite{khazoom_tailoring_2025}, and the schedule-design framework of~\cite{khazoom_optimal_2023} is restricted to offline settings.
In parallel, recent results suggest that tailored \gls{sqp} solvers are highly competitive for constrained nonlinear \gls{mpc}~\cite{jordana_structure-exploiting_2025} and are available in mature general-purpose frameworks such as acados~\cite{verschueren_acadosmodular_2022}, which has direct support for multi-phase problems~\cite{frey_multi-phase_2025}.

\begin{figure}[t]
    \includegraphics[width=\linewidth]{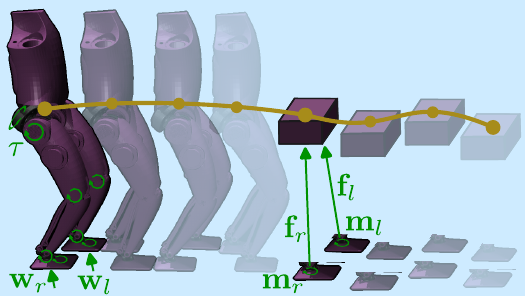}
    \vspace{-1.8em}
    \caption{The multi-phase \acrfull{mpc} predicts the first \(N_\mathrm{wb}\) steps using the \acrfull{wb} dynamics, while the following \(N_\mathrm{srb}\) steps are predicted using \acrfull{srb} dynamics.  This way prediction accuracy and solve time can be balanced effectively.}
    \vspace{-1.8em}
    \label{fig:multi_phase_mpc}
\end{figure}

Motivated by these developments, this paper presents a whole-body multi-phase \gls{mpc} for bipedal walking. The proposed controller employs a \gls{wb} model in the early steps of the horizon and a \gls{srb} model in the later steps (see Fig. \ref{fig:multi_phase_mpc}) to reduce computational cost, while solving the resulting nonlinear optimal control problem in a real-time \gls{sqp} setting with only a small number of iterations per control cycle.
In contrast to prior multi-phase MPC works~\cite{li_model_2021, li_cafe-mpc_2025, khazoom_tailoring_2025}, which rely on custom \gls{ddp} solvers or hand-built \gls{sqp} layers, the entire formulation is implemented in acados~\cite{verschueren_acadosmodular_2022}, a general-purpose nonlinear MPC framework. 
We additionally provide an empirical characterization of how the WB/SRB split, discretization steps, and number of \gls{sqp} iterations jointly affect real-time feasibility and tracking accuracy.
Our controller was successfully tested on the \num{18} \gls{dof} bipedal robot \textit{HyPer-2} in a MuJoCo simulation using purely torque control and a control frequency of \qty{100}{\hertz}.

\section{Model Predictive Control Formulation}
The proposed \gls{mpc} scheme is formulated as the following two-phase optimal control problem, in which a short-horizon \gls{wb} model is followed by a reduced \gls{srb} model.
\begin{mini}|s|[1]
{\mathcal{X},\mathcal{U}}
{%
\begin{aligned}
&\sum_{k=0}^{N_{\mathrm{wb}}-1}
L_\mathrm{wb}\!\left(\mathbf{x}_{\mathrm{wb},k}, \mathbf{u}_{\mathrm{wb},k}\right)
+ L_\mathrm{t}\!\left(\mathbf{x}_{\mathrm{wb},N_{\mathrm{wb}}}\right) \\
&\quad
+ \sum_{k=0}^{N_{\mathrm{srb}}-1}
L_\mathrm{srb}\!\left(\mathbf{x}_{\mathrm{srb},k}, \mathbf{u}_{\mathrm{srb},k}\right)
+ L_\mathrm{f}\!\left(\mathbf{x}_{\mathrm{srb},N_{\mathrm{srb}}}\right)
\end{aligned}
}
{\label{eq:mpc-form}}{}
\addConstraint{\mathbf{x}_{\mathrm{wb}, 0}}{= \overline{\mathbf{x}}}
\addConstraint{\mathbf{x}_{\mathrm{wb},k+1}}{= f_\mathrm{wb}\left(\mathbf{x}_{\mathrm{wb},k}, \mathbf{u}_{\mathrm{wb},k}\right) \quad && k=0\dots N_\mathrm{wb}-1}
\addConstraint{0}{\geq h_\mathrm{wb}\left(\mathbf{x}_{\mathrm{wb},k}, \mathbf{u}_{\mathrm{wb},k}\right) \quad && k=0\dots N_\mathrm{wb}-1}
\addConstraint{0}{\geq h_{\mathrm{t}}\left(\mathbf{x}_{\mathrm{wb},N_\mathrm{wb}}\right)}
\addConstraint{\mathbf{x}_{\mathrm{srb},0}}{= f_\mathrm{t}\left(\mathbf{x}_{\mathrm{wb},N_\mathrm{wb}}\right)}
\addConstraint{\mathbf{x}_{\mathrm{srb},k+1}}{= f_\mathrm{srb}\left(\mathbf{x}_{\mathrm{srb},k}, \mathbf{u}_{\mathrm{srb},k}\right) \quad && k=0\dots N_\mathrm{srb}-1}
\addConstraint{0}{\geq h_\mathrm{srb}\left(\mathbf{x}_{\mathrm{srb},k}, \mathbf{u}_{\mathrm{srb},k}\right) \quad && k=0\dots N_\mathrm{srb}-1}
\addConstraint{0}{\geq h_{\mathrm{f}}\left(\mathbf{x}_{\mathrm{srb},N_\mathrm{srb}}\right)}.
\end{mini}
Here, \(\mathcal{X}\) and \(\mathcal{U}\) denote the set of all states and inputs over both phases. 
The cost function is composed of the stage costs of the \gls{wb} and \gls{srb} phases \(L_\mathrm{wb}(\cdot)\) and \(L_\mathrm{srb}(\cdot)\), a transition cost at the phase boundary \(L_\mathrm{t}(\cdot)\), and a terminal cost at the end of the prediction horizon \(L_\mathrm{f}(\cdot)\). 
The first phase starts from the current state estimate \(\bar{\mathbf{x}}\) and is subject to the \gls{wb} dynamics \(f_\mathrm{wb}(\cdot)\) and constraints \(h_\mathrm{wb}(\cdot)\). At the switching point, the transition map \(f_\mathrm{t}(\cdot)\) initializes the \gls{srb} state from the terminal \gls{wb} state, while additional transition constraints are imposed through \(h_\mathrm{t}(\cdot)\). The second phase then evolves according to the \gls{srb} dynamics  \(f_\mathrm{srb}(\cdot)\), constraints  \(h_\mathrm{srb}(\cdot)\) and  the terminal constraint set \(h_\mathrm{f}(\cdot)\).
All terms are detailed in the following.
\subsection{Whole Body phase}
The \gls{wb}-phase considers the full bipedal dynamics. 
The state is parameterized as
\(\mathbf{x}_{\mathrm{wb},k} = [\delta \mathbf{q}_k^\top, \mathbf{v}_k^\top]^\top\), where \(\delta \mathbf{q}_k \in \mathbb{R}^{n_j + 6}\) denotes a local coordinate representation of the generalized positions around the initial generalized position \(\mathbf{q_0} \in \mathbb{R}^{n_j+7}\), with \(n_j\) the number of actuated joints, the floating-base orientation expressed in the tangent space of \(SO(3)\), and \(\mathbf{v}_k \in \mathbb{R}^{n_j + 6}\) the generalized joint velocities.
The full configuration for prediction step $k$ can be recovered as \(\mathbf{q}_k = \mathbf{q_0} \oplus \delta \mathbf{q}_k\),
where \(\oplus\) denotes the manifold update.
The control input is defined as
\(\mathbf{u}_{\mathrm{wb},k} = [\boldsymbol{\tau}_k^\top, \mathbf{w}_{\mathrm{wb},\mathrm{l},k}^\top, \mathbf{w}_{\mathrm{wb},\mathrm{r},k}^\top]^ \top \in  \mathbb{R}^{n_j + 12}\),
where \(\boldsymbol{\tau} \in \mathbb{R}^{n_j}\) denotes the vector of actuated joint torques and
\(\mathbf{w}_{\mathrm{wb},i,k} = \left[\mathbf{f}_{\mathrm{wb},i,k}^\top, \mathbf{m}_{\mathrm{wb},i,k}^\top \right]^\top \in \mathbb{R}^{6}\) the contact wrenches for the left and right foot (\(i \in \left\{\mathrm{l}, \mathrm{r}\right\}\)), composed of the contact force  \(\mathbf{f}_{\mathrm{wb},i,k} \in \mathbb{R}^3\) and contact moment \(\mathbf{m}_{\mathrm{wb},i,k} \in \mathbb{R}^3\).
The system dynamics are discretized using forward Euler, yielding
\begin{equation}
    f_\mathrm{wb}\left(\mathbf{x}_{\mathrm{wb},k}, \mathbf{u}_{\mathrm{wb},k}\right) = \begin{bmatrix}\delta \mathbf{q}_{k+1} \\
    \mathbf{v}_{k+1}
    \end{bmatrix} =
    \begin{bmatrix}\delta \mathbf{q}_{k} + \Delta t_\mathrm{wb}\, \mathbf{v}_{k} \\
    \mathbf{v}_{k} + \Delta t_\mathrm{wb} \, \mathbf{a}_{k} 
    \end{bmatrix}
\end{equation}
where \(\mathbf{a}_k\) denotes the generalized acceleration and \(\Delta t_\mathrm{wb}\) the discretization step. 
To reduce computational complexity, dynamics terms are fixed around the initial  \((\mathbf{q_0}, \mathbf{v_0})\) such that
\begin{equation}
\small
    \mathbf{a}_{k} = \mathbf M(\mathbf{q}_0)^{-1}
\left(
\mathbf B \boldsymbol{\tau}_k
-
\mathbf n(\mathbf{q}_0,\mathbf{v}_0)
+ \sum_{i\in\left\{\mathrm{l}, \mathrm{r}\right\}}\mathbf J_{c_i}(\mathbf q_0)^\top \mathbf w_{\mathrm{wb},i,k}
\right)
,
\end{equation}
where \(\mathbf M(\mathbf q_0) \in \mathbb{R}^{(n_j + 6) \times (n_j + 6)}\) is the generalized inertia matrix, \(\mathbf B \in \mathbb{R}^{(n_j + 6) \times n_j}\) is the actuation selection matrix, \(\mathbf J_{c_i}(\mathbf q_0) \in \mathbb{R}^{6 \times (n_j + 6)}\) is the Jacobian of contact \(i\), and \(\mathbf n(\mathbf q_0,\mathbf v_0) \in \mathbb{R}^{n_j + 6}\) is the combined vector of Coriolis-Centrifugal and gravity forces.
The \gls{wb} phase is augmented with the following constraints, which together define \(h_\mathrm{wb}(\cdot)\):
\subsubsection{Wrench cone}
Each \(\mathbf{w}_{\mathrm{wb},i, k}\) is constrained to a rectangular wrench cone as derived in \cite{caron_stability_2015}.
\subsubsection{Contact schedule}
The force-contact constraints couple the normal contact force to the prescribed contact mode by enforcing \(0 \leq \left[\mathbf{f}_{\mathrm{wb},i,k}\right]_\mathrm{z} \leq c_{\mathrm{wb},i,k} \, f_z^{\max}\), where \(c_{\mathrm{wb},i,k} \in \{0,1\}\) is prescribed a priori and indicates whether foot \(i\) is planned to be in contact at step \(k\), \([\cdot]_z\) extracts the \(z\)-component of the respective force and \( f_z^{\max}\) is the maximal z-force.
\subsubsection{No slip}
The no-slip constraints enforce zero foot velocity for active contacts, i.e.,
\(c_{\mathrm{wb},i,k}\,\mathbf{J}_{c_i}(\mathbf q_0)\,\mathbf v_k = 0\).
\subsubsection{Contact kinematics}
For each foot \(i\), a ground-contact consistency constraint is imposed as
\begin{align}
0 &\le h_i(\mathbf q_k) \le (1-c_{\mathrm{wb},i,k})\,\mathrm{h}_{\max},\\
c_{\mathrm{wb},i,k}\,\phi_i(\mathbf q_k) &= 0,\qquad
c_{\mathrm{wb},i,k}\,\theta_i(\mathbf q_k) = 0,
\end{align}
where the functions \(h_i(\cdot)\), \(\phi_i(\cdot)\), and \(\theta_i(\cdot)\) map the generalized joint positions to the foot height, foot roll and foot pitch angle of foot \(i\in\{\mathrm{l}, \mathrm{r}\}\) using forward kinematics.
This enforces zero height and planar contact under the assumption of planar terrain, while during swing allowing a maximum height of \(\mathrm{h}_{\max}\).
\subsubsection{Joint limits}
Joint position, velocity and torque limits are enforced as box constraints.

The cost function of the \gls{wb}-phase is
\begin{align}\label{eq:cost-wb}
    L_\mathrm{wb}(\cdot) &=
    \left\|\delta \mathbf{q}_k - \delta \mathbf{q}_\mathrm{ref}\right\|^2_{\mathrm{\mathbf{W}_q}}
    + \left\|\mathbf{v}_k - \mathbf{v}_\mathrm{ref}\right\|^2_{\mathrm{\mathbf{W}_v}}\\+ & \sum_{i\in\left\{\mathrm{l},\mathrm{r}\right\}}{\left\|h_{i}\left(\mathbf{q}_k\right)- h_{i,k,\mathrm{ref}} \right\|^2_\mathrm{w_h}} + \left\|\phi_i(\mathbf q_k)\right\|^2_\mathrm{w_\phi} \nonumber\\
    & \qquad +\left\|\theta_i(\mathbf q_k)\right\|^2_\mathrm{w_\theta} +\left\|{}^\mathrm{b} \psi_i(\mathbf q_k)\right\|^2_\mathrm{{w_\psi}}
    \nonumber\\
    & \qquad + \|{}^\mathrm{b} y_\mathrm{i, dis}\left(\mathbf{q}_k\right)\|^2_{\mathrm{w_{yb}}} + \left\| \mathbf{\dot{ p}}_i\left(\mathbf{q}_k,\mathbf{v}_k\right) -  \mathbf{\dot p}_{i,\mathrm{ref}} \right\|^2_\mathrm{\mathbf{W}_p} \nonumber\\
    + & \left\|\mathbf{u}_{\mathrm{wb},k}\right\|^2_\mathrm{\mathbf{W}_u} \nonumber,
\end{align}
where \(\delta \mathbf{q}_\mathrm{ref}\), \(\mathbf{v}_\mathrm{ref}\) denote the local configuration increment and generalized velocity references, respectively. Furthermore \(h_{i,\mathrm{ref}}\) denotes the height reference of foot \(i\) at prediction step \(k\), 
\({}^\mathrm{b} \psi_i(\mathbf q_k)\) denotes the foot yaw in body coordinates obtained via forward kinematics.
Further, \({}^\mathrm{b} y_\mathrm{i, dis}\left(\mathbf{q}_k\right)\) is the foot–hip distance along the body \(y\)-axis of foot \(i\), respectively.
Finally,  \(\dot{\mathbf p}_i(\mathbf q_k,\mathbf v_k)\) denotes the Cartesian foot velocity with \(\mathbf{\dot p}_{i,\mathrm{ref}}\) being its reference. 
The diagonal weighting matrices and scalars \(\mathrm{\mathbf{W}_q}\), \(\mathrm{\mathbf{W}_v}\), \(\mathrm{w_h}\), \(\mathrm{w_\phi}\), \(\mathrm{w_\theta}\), \(\mathrm{w_\psi}\), \(\mathrm{w_{yb}}\), \(\mathrm{\mathbf{W}_p}\), and \(\mathrm{\mathbf{W}_u}\) determine the relative importance of the individual objectives.
While the dynamic constraint is linear in the decision variables, the overall \gls{wb}-phase remains nonlinear due to the kinematic constraints and the forward-kinematics-dependent cost terms, which are evaluated at the predicted \(\mathbf{q}_k\).

\subsection{Single Rigid Body phase}
In the \gls{srb} phase, the robot is approximated as a single rigid body with lumped mass and inertia, while the contact forces and contact locations are optimized.
The state is parameterized as \(\mathbf{x}_{\mathrm{srb},k} = [\delta \mathbf{q}_{\mathrm{b}, k}^\top, \mathbf{v}_{\mathrm{b}, k}^\top, \mathbf{p}_{\mathrm{l},k}^\top, \mathbf{p}_{\mathrm{r},k}^\top]^\top\), where \(\delta \mathbf{q}_{\mathrm{b}, k} \in \mathbb{R}^6\) and \(\mathbf{v}_{\mathrm{b}, k} \in \mathbb{R}^6 \) denote local coordinate representation of the floating base position and the velocity. Hence they match the parameterization of the \gls{wb}-phase state such that the floating base position \(\mathbf{q}_{\mathrm{b}, k} \) can be recovered from \(\mathbf{q}_0\) as in the \gls{wb}-phase. 
Instead of capturing the further joints in the state, the foot positions relative to the body frame are directly included as \(\mathbf{p}_{i,k} \in \mathbb{R}^3\).
The control input is given by \(\mathbf{u}_{\mathrm{srb},k} = [\mathbf{w}_{\mathrm{srb}, \mathrm{l},k}^\top,\mathbf{w}_{\mathrm{srb},\mathrm{r},k}^\top,\dot{\mathbf{p}}_{\mathrm{l},k}^\top,\dot{\mathbf{p}}_{\mathrm{r},k}^\top]^\top\),
with \(\dot{\mathbf{p}}_{i,k}\) as the foot velocities in body and the wrenches analogue to \gls{wb}-phase as \(\mathbf{w}_{\mathrm{srb},i,k} = [\mathbf{f}_{\mathrm{srb},i,k}^\top,\mathbf{m}_{\mathrm{srb},i,k}^\top]^\top \in \mathbb{R}^6\).
The state update is
\begin{align}
f_\mathrm{srb}\left(\cdot\right) =
\begin{bmatrix}
\delta \mathbf q_{\mathrm{b},k+ 1} \\
\mathbf v_{\mathrm{b},k+1}\\
\mathbf{p}_{\mathrm{l},k+1}\\
\mathbf{p}_{\mathrm{r},k+1}
\end{bmatrix} =
\begin{bmatrix}
\delta \mathbf{q}_{\mathrm{b},k}+ \Delta t_\mathrm{srb} \mathbf v_{b,k} \\
\mathbf{v}_{\mathrm{b},k}+ \Delta t_\mathrm{srb} \mathbf{a}_{\mathrm{b},k} \\
\mathbf{p}_{\mathrm{l},k}+ \Delta t_\mathrm{srb} \dot{ \mathbf{p}}_{\mathrm{l},k} \\
\mathbf{p}_{\mathrm{r},k}+ \Delta t_\mathrm{srb} 
\dot{\mathbf{p}}_{\mathrm{r},k} 
\end{bmatrix},
\end{align}
with the acceleration being expressed in local robot frame
\begin{align}
\small
   \mathbf{a}_{\mathrm{b},k} = \begin{bmatrix}
   \frac{1}{\mathrm{m}}\left(\sum_{i\in\left\{\mathrm{l},\mathrm{r}\right\}}\mathbf{f}_{\mathrm{srb},i,k}\right) - \mathbf R^\mathrm{w}_\mathrm{b}(\mathbf q_0)\,\mathrm{\mathbf{g}}
   \\
   \mathbf{I}\left(\mathbf{q_0}\right)^{-1}\left(\sum\limits_{i\in\left\{\mathrm{l},\mathrm{r}\right\}}\left[\mathbf{p}_{i,k}- \mathbf{p}_{\mathrm{c}}\left(\mathbf{q_0}\right)\right]\times \mathbf{f}_{\mathrm{srb}, i,k} + \mathbf{m}_{\mathrm{srb},i, k}\right)
   \end{bmatrix}
\end{align}
\(m\) is the total robot mass, \(\mathbf{g}\) is the gravity vector transformed into body frame by \(\mathbf{R}^\mathrm{w}_\mathrm{b}(\mathbf q_0)\), \(\mathbf{I}(\mathbf{q}_0)\) is the rigid-body inertia used in the \gls{srb} phase, and \(\mathbf{p}_{\mathrm{c}}(\mathbf{q}_0)\) denotes the \gls{com} position relative to the body frame, both linearized around \(\mathbf{q}_0\).
The \gls{srb}-phase is subject to the following constraints, which together define \(h_\mathrm{srb}(\cdot)\) and \(h_\mathrm{f}(\cdot)\):

\subsubsection{Kinematic reachability}
Each foot is restricted to a kinematically admissible region. Since the relative foot positions are states of the \gls{srb} model, this is imposed directly as
\begin{align}
\underline{\mathbf{p}}_i \leq \mathbf{p}_{i,k} \leq \overline{\mathbf{p}}_i, \qquad \text{for } i\in\left\{\mathrm{l},\mathrm{r}\right\},
\end{align}where the lateral bounds are chosen side-dependently to preserve the left/right foot assignment.
\subsubsection{Wrench cones}
As in the \gls{wb} phase, all contact wrenches \(\mathbf{w}_{\mathrm{srb},i,k}\) are constrained into rectangular wrench cones.
\subsubsection{Contact schedule}
The normal contact forces are coupled to the contact schedule as in the \gls{wb}-phase.

\subsubsection{Contact kinematics}
Ground consistency is enforced as
\begin{equation}
0 \leq \left[ {}^\mathrm{w} \mathbf{p}_{i,k}\right]_\mathrm{z}  \leq (1 - c_{\mathrm{srb},i,k}) \, \mathrm{h}_{\max} , \qquad i \in \left\{\mathrm{l}, \mathrm{r}\right\},
\end{equation}
where \(c_{\mathrm{srb},i,k}\) specifies the contact schedule as in \gls{wb}-phase and \([\cdot]_z\) extracts the \(z\)-component of the respective foot position approximated into world coordinates by
\begin{align}
   {}^\mathrm{w} \mathbf{p}_{i,k} = \left[\mathbf{q}_{\mathrm{b},k}\right]_\mathrm{trans} + \mathbf{R}^\mathrm{w}_\mathrm{b}(\mathbf q_0) \mathbf{p}_{i,k}.
\end{align}
Here, \([\cdot]_{\mathrm{trans}}\) extracts the translational part of the floating-base generalized positions or velocities.

\subsubsection{No slip}
The no-slip condition in the \gls{srb} phase can be directly imposed through foot velocities,
\begin{align}
c_{\mathrm{srb},i,k}\left(\left[\mathbf{v}_{\mathrm{b},k}\right]_\mathrm{trans} + \dot{\mathbf p}_{i,k}\right) = \mathbf 0,
\qquad i\in\{\mathrm{l},\mathrm{r}\}.
\end{align}

The cost function of the \gls{srb}-phase is
\begin{align}\label{eq:cost-srb}    L_\mathrm{srb}\left(\cdot\right) &= 
    \left\|\mathbf{\delta q_{\mathrm{b},k}} - \delta \mathbf{q}_{\mathrm{b}, \mathrm{ref}}\right\|^2_\mathrm{\mathbf{W}_{qb}} 
    + \left\|\mathbf{v_{\mathrm{b},k}} - \mathbf{v}_{\mathrm{b}, \mathrm{ref}} \right \|^2_\mathrm{\mathbf{W}_{vb}}\nonumber\\
    +&\sum_{i\in\left\{\mathrm{l}, \mathrm{r}\right\}}
    \left\|\mathbf{p}_{i,k} - \mathbf{p}_{i,\mathrm{ref}}\right\|^2_\mathrm{\mathbf{W}_{p}}
    + \left\|\left[{}^\mathrm{w} \mathbf{p}_{i,k}\right]_\mathrm{z} - \mathrm{h}_{i,k,\mathrm{ref}}\right\|^2_\mathrm{w_\mathrm{h}} \nonumber
    \\ & + \left\|\mathbf{u}_{\mathrm{srb},k}\right\|^2_\mathrm{\mathbf{W}_{u\mathrm{srb}}},
\end{align}
where \(\delta \mathbf{q}_{\mathrm{b},\mathrm{ref}}\), \(\mathbf{v}_{\mathrm{b},\mathrm{ref}}\) denote the local floating-base configuration increment and base velocity references, respectively. Furthermore, \(\mathbf{p}_{i,\mathrm{ref}}\) is the relative position reference of foot \(i\in\{\mathrm{l},\mathrm{r}\}\) in the body frame. The weighting matrices and scalars \(\mathrm{\mathbf{W}_{qb}}\), \(\mathrm{\mathbf{W}_{vb}}\), \(\mathrm{\mathbf{W}_{p}}\), \(\mathrm{w_{\mathrm h}}\), and \(\mathrm{\mathbf{W}_{u\mathrm{srb}}}\) determine the relative importance of the individual objectives. 
The final cost function \(L_\mathrm{f}(\cdot)\) uses the same terms, except the input dependent.
\subsection{Transition phase}
The transition-phase maps the \gls{wb}-state to the \gls{srb}-state as
\begin{equation}
    f_\mathrm{t}(\cdot) = \begin{bmatrix}
\delta \mathbf q_{\mathrm{b},0} \\
\mathbf v_{\mathrm{b},0}\\
\mathbf{p}_{\mathrm{l},0}\\
\mathbf{p}_{\mathrm{r},0}
\end{bmatrix}=
\begin{bmatrix}
\left[\delta \mathbf{q}_{N_\mathrm{wb}}\right]_\mathrm{fb} \\
\left[\mathbf{v}_{N_\mathrm{wb}}\right]_\mathrm{fb} \\
{}^b p_\mathrm{l}\!\left(\mathbf{\mathbf{q}}_{N_\mathrm{wb}}\right)\\
{}^b p_\mathrm{r}\!\left(\mathbf{\mathbf{q}}_{N_\mathrm{wb}}\right)
\end{bmatrix},
\end{equation}
where \([\cdot]_{\mathrm{fb}}\) extracts the floating-base components (first 6 elements) of the corresponding whole-body state, and \({}^b p_i\!\left(\mathbf q_{N_\mathrm{wb}}\right)\), \(i\in\{\mathrm{l},\mathrm{r}\}\), denotes the foot position in body frame obtained from the forward kinematics at the terminal whole-body configuration.
The final state of the \gls{wb}-phase \(\mathbf{x}_{\mathrm{wb}, {N_\mathrm{wb}}}\) is constrained and penalized in the transition phase by \(h_\mathrm{t}(\cdot)\) and \(L_\mathrm{t}(\cdot)\), which both assemble from the same components as  \(h_\mathrm{wb}(\cdot)\) and \(L_\mathrm{wb}(\cdot)\), omitting the input dependent terms.
\section{Results}
The controller is evaluated in a MuJoCo simulation on the torque controlled bipedal robot \textit{HyPer-2}, developed at DFKI RIC.
The robot has a height of \qty{110}{\cm}, weighs \qty{22.00}{\kg} and has \num{6} torque-controlled joints per leg. In this work it is considered without arms and hence has in total \num{18} \glspl{dof} including  the floating base.
The robot dynamics and kinematics are generated symbolically with Pinocchio~\cite{carpentier2019pinocchio} and CasADi~\cite{Andersson2019}, and used to formulate the MPC in acados~\cite{verschueren_acadosmodular_2022}.
Initially, the \gls{mpc} is manually tuned to achieve stable walking. Therefore, the ratio \(\alpha = \frac{N_\mathrm{wb}}{N_\mathrm{wb} + N_\mathrm{srb}}\) is set to \(\alpha=0.5\) and the total number of prediction steps to \(N=N_\mathrm{wb} + N_\mathrm{srb}=10\). It was observed that increasing the discretization step size along the prediction horizon is beneficial. Hence the step sizes were set to \(\Delta t_\mathrm{wb}=\qty{0.02}{\second},\, \Delta t_\mathrm{srb}=\qty{0.1}{\second}\). The weights in \eqref{eq:cost-wb} and \eqref{eq:cost-srb} are tuned  so that the robot maintains a fixed target height and follows a fixed target speed (see Appendix \ref{ap:tunings}). A simple walking gait with a per-foot stance time of \qty{0.5}{\second} and a double support time of \qty{0.1}{\second} is used.
The swing-foot height is generated as a smooth arc with \qty{3}{\centi\meter} clearance, whose derivative defines vertical velocity targets that are zero at lift-off and touchdown.
Stable walking under pure torque control required a \qtyrange{50}{100}{\hertz} control rate, leaving a maximum solve time of \qty{10}{\milli\second}.
The solver parameters, shown in Appendix~\ref{sec:solver_settings}, were manually tuned to ensure stability while also finding solutions quickly. 
\begin{figure}[t]
    \centering
    \includegraphics{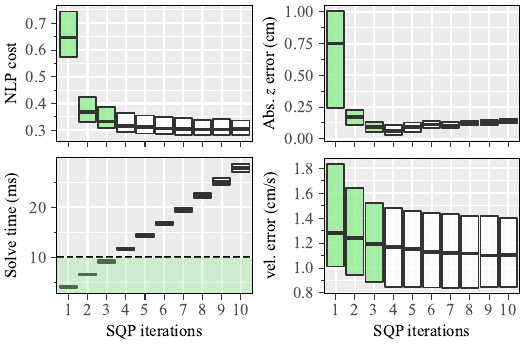}
    \vspace{-2.0em}
    \caption{Average NLP cost, solve time, height and velocity error for bipedal walking over no. of \gls{sqp} iterations. RT capable settings are marked in green.}
    \vspace{-1.2em}
    \label{fig:sqp_iters}
\end{figure}
However, as shown in \cite{khazoom_optimal_2023}, it is not strictly necessary to solve the optimization problem until convergence, since a ``good-enough'' solution is usually adequate. Consequently, the number of \gls{sqp} iterations remains one of the main factors influencing the solve time. To investigate this further, we repeat the following experiment several times: The bipedal robot walks forward for \qty{5}{\second} at a target speed of \qty{0.3} meters per second. We increase the number of \gls{sqp} iterations in each run, starting at 1. The control frequency was set to \qty{100}{\hertz}, regardless of the solve time. \autoref{fig:sqp_iters} shows the average solve time, \gls{nlp} cost, height and velocity error over the maximum number of \gls{sqp}-iterations, respectively.
\begin{figure}[h]
    \centering
    \vspace{-0.2em}
    \includegraphics{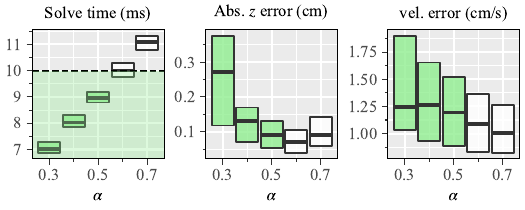}
    \vspace{-2.2em}
    \caption{Average solve time, height and velocity error for bipedal walking. Here, the ratio \(\alpha = N_\mathrm{wb}/(N_\mathrm{wb} + N_\mathrm{srb})\) is varied, while \(N\) is kept fixed ($N=10$). RT capable settings are marked in green.}
    \vspace{-0.2em}
    \label{fig:alphas}
\end{figure}
It shows that solve time is smaller than \qty{10}{\milli\second} when using up to \num{3} \gls{sqp} iterations and the solve time increases linearly with the number of iterations. The height and velocity errors are largely minimized within the first 3 iterations, while the solver cost does not change much after the first 3 iterations. Hence, \num{3} was chosen as the iteration limit for the controller and all further experiments.
A video demonstrating the simulated \textit{HyPer-2} robot walking under real-time torque control is available online\footnote{Video of the robot walking in simulation: \url{https://youtu.be/1VuRR2P5hxM}}.
To investigate the relationship between the prediction horizon of whole-body and single rigid body model, the ratio \(\alpha\) is varied, while keeping $N=10$ fixed. The results are shown in \autoref{fig:alphas}. When using $N_{wb} \leq 2$, that is \(\alpha \in \left\{0.1, 0.2 \right\}\), the robot quickly became unstable. When increasing $N_{wb}$, both, the height and velocity error reduce, where the former has its minimum at \(\alpha=0.6\). At the same time, the solve time increases linearly, such that for \(\alpha=0.6\) the target solve time of \qty{10}{\milli\second} is not met. For \(\alpha > 0.7\), the robot fell.
\begin{figure}[t]
     \vspace{-0.3em}
    \centering
    \includegraphics{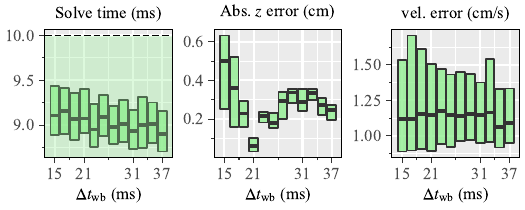}
    \vspace{-2.5em}
    \caption{Average solve time, height and velocity error for bipedal walking. Here, \(\Delta t_\mathrm{wb}\) is varied and \(\Delta t_\mathrm{srb} = 5\,\Delta t_\mathrm{wb}\).}
    \vspace{-0.5em}
    \label{fig:dts}
\end{figure}
Finally, we vary the discretization step sizes to study the influence of the resulting prediction horizon length, \(
T = N_\mathrm{wb}\,\Delta t_\mathrm{wb} + N_\mathrm{srb}\,\Delta t_\mathrm{srb}. \)
Here, \(\Delta t_\mathrm{wb}\) was varied directly, while \(\Delta t_\mathrm{srb}\) was computed according to \(\Delta t_\mathrm{srb} = 5\,\Delta t_\mathrm{wb}\).
The results are shown in \autoref{fig:dts}. Unlike in the two previous experiments, the discretization step sizes only have a minor influence on the solve time. The robot is unstable for \(\Delta t_\mathrm{wb} < \qty{15}{\milli\second}\) (\(T < \qty{90}{\milli\second}\)).
While the velocity error is largely independent of the discretization step size, the height error is smallest at approximately \(\Delta t_\mathrm{wb} = \qty{21}{\milli\second}\) (\(T = \qty{126}{\milli\second}\)).
Beyond a step size of \(\qty{21}{\milli\second}\), the robot exhibits increasing knee buckling and approaches a singular configuration, eventually falling for step sizes above \(\Delta t_\mathrm{wb} > \qty{37}{\milli\second}\) (\(T > \qty{222}{\milli\second}\)).

\section{Discussion and Future Work}
The results demonstrate that real-time whole-body \gls{mpc} with a multi-phase horizon can be achieved within a general-purpose, off-the-shelf nonlinear \gls{mpc} framework, without requiring a custom solver tailored to legged robotics. This lowers the implementation barrier and allows direct benefit from ongoing advances in numerical optimization. 
Stable walking is obtained with only a small number of \gls{sqp} iterations, and the reduced model complexity in the further part of the prediction horizon decreases solve time effectively.
However, the analysis is not yet conclusive. Varying the solve time in reality degrades control frequency.
The horizon composition via \(\alpha\) also changes the effective prediction horizon \(T\) due to the different discretization step sizes in each phase. As a result, the instabilities we observe for larger values of \(\alpha\) (e.g., \(\alpha > 0.7\)) may also be attributed to a shortened horizon. 

Future work will therefore focus on resolving this through more controlled experiments that independently vary horizon length and model composition. 
Extending the framework to further model abstractions, and relaxing the prescribed contact schedule are other promising directions. Finally, validation on the real \textit{HyPer-2} system is required to assess stability under realistic conditions.

\newpage
\section*{Acknowledgements}
This work was done in the projects CoEx (grant number 01IW24008) and ActGPT (grant number 01IW25002) both funded by the Federal Ministry of Research, Technology and Space (BMFTR) and is supported with funds from the federal state of Bremen for setting up the Underactuated Robotics Lab (265/004-08-02-02-30365/2024-102966/2024-740847/2024).
Further, this work has been partially supported by the German Federal Ministry of Research, Technology and Space (BMFTR) under the Robotics Institute Germany (RIG).
Many thanks to the acados team, for enabling such results.

\appendix
\subsection{Cost function weights}\label{ap:tunings}
Table~\ref{tab:tunings} summarizes the diagonal entries of all weighting matrices and scalars used in the \gls{wb} and \gls{srb} stage and terminal costs. In the \gls{wb} phase, the foot $z$-velocity costs are contact-stage dependent. 
During swing, moderate costs are used, while in approach (one timestep before contact), larger weights promote a controlled touchdown. The foot-hip distance regularization enforces more natural walking and also works against foot self-collisions. 
In the \gls{srb} phase, foot position weights enforce natural below-hip foot placement.
\begin{table}[h]
\vspace{-1.5em}
\centering
\caption{Diagonal cost weights used in the WB and SRB phases.}
\label{tab:tunings}
\scriptsize
\setlength{\tabcolsep}{4pt}
\begin{tabular}{lll}
\toprule
\textbf{Symbol} & \textbf{Description} & \textbf{Value} \\
\midrule
\multicolumn{3}{l}{\textit{WB phase}}\\
\midrule
\(\mathbf{W}_\mathrm{q}\) 
& base pos./ori. \((x,y,z,\phi,\theta,\psi)\) 
& \((0,0,50,10,10,1)\) \\
& joint position regularization
& \(1\times10^{-8}\) \\

\(\mathbf{W}_\mathrm{v}\) 
& base vel. \((v_x,v_y,v_z,\omega_x,\omega_y,\omega_z)\) 
& \((40,40,10,10,10,1)\) \\
& joint velocity 
& \(0.05\) \\

\(w_\mathrm{h}\) 
& foot height 
& \(10\) \\
\(w_\mathrm{\phi},\,w_\mathrm{\theta}\) 
& foot roll/pitch 
& \(30\) \\
\(w_\mathrm{\psi}\) 
& foot yaw 
& \(60\) \\
\(w_\mathrm{yb}\) 
& foot--hip distance in body \(y\) 
& \(10\) \\

\(\mathbf{W}_\mathrm{p}\) 
& foot velocity (swing) 
& \((1,1,1)\) \\
& foot velocity (approach) 
& \((10,10,30)\) \\

\(\mathbf{W}_\mathrm{u}\) 
& joint torque regularization 
& \(1\times10^{-13}\) \\
& contact wrench regularization 
& \(1\times10^{-8}\) \\
\midrule
\multicolumn{3}{l}{\textit{SRB phase}}\\
\midrule
\(\mathbf{W}_\mathrm{qb}\) 
& base pos./ori. \((x,y,z,\phi,\theta,\psi)\) 
& \((0,0,50,10,10,1)\) \\
\(\mathbf{W}_\mathrm{vb}\) 
& base vel. \((v_x,v_y,v_z,\omega_x,\omega_y,\omega_z)\) 
& \((40,40,10,10,10,1)\) \\
\(\mathbf{W}_\mathrm{p}\) 
& foot position in body \(xy\) 
& \(100\) \\
\(w_\mathrm{h}\) 
& foot height 
& \(10\) \\
\(\mathbf{W}_\mathrm{p}\) 
& foot velocity 
& \(2\) \\
\(\mathbf{W}_\mathrm{u,srb}\) 
& contact wrench regularization 
& \(1\times10^{-8}\) \\
\bottomrule
\end{tabular}
 \vspace{-1.5em}
\end{table}
\subsection{Solver setup}\label{sec:solver_settings}
The acados \gls{sqp} solver is configured to use a Gauss-Newton Hessian approximation and the sparsity exploiting \gls{ipm} solver HPIPM \cite{frison_hpipm_2020} in \texttt{BALANCE} for the \gls{qp}-subproblems. Warm-starting is enabled at both the \gls{nlp} and \gls{qp} levels, including initialization of the first \gls{qp} from the previous \gls{nlp} solution. An adaptive Levenberg-Marquardt regularization is employed to improve numerical robustness, together with projection-based regularization of the \gls{qp}.
A fixed-step globalization strategy is used, avoiding line-search overhead and favoring predictable runtimes. 
The adaptive \gls{qp} tolerance strategy \texttt{ADAPTIVE\_CURRENT\_RES\_JOINT} balances accuracy and speed during each \gls{sqp} iteration.

\bibliographystyle{IEEEtran}
\bibliography{references} 

\end{document}